\pdfoutput=1

\documentclass[11pt]{article}

\usepackage{acl}

\usepackage{times}
\usepackage{latexsym}
\usepackage{amsmath}
\usepackage{amssymb}
\usepackage[T1]{fontenc}
\usepackage[utf8]{inputenc}
\usepackage{microtype}
\usepackage{inconsolata}
\usepackage{graphicx}
\usepackage{algorithm}
\usepackage{algorithmic}
\usepackage{booktabs}
\usepackage{float}

\DeclareMathOperator*{\argmax}{arg\,max}

\title{Rationale-Aware Answer Verification by Pairwise Self-Evaluation}

\author{Akira Kawabata \\
  The Asahi Shimbun Company \\
  \texttt{kawabata-a@asahi.com} \\\And
  Saku Sugawara \\
  National Institute of Informatics \\
  \texttt{saku@nii.ac.jp} \\}

\begin{document}
\maketitle
\begin{abstract}

Answer verification identifies correct solutions among candidates generated by large language models (LLMs). 
Current approaches typically train verifier models by labeling solutions as correct or incorrect based solely on whether the final answer matches the gold answer.
However, this approach neglects any flawed rationale in the solution yielding the correct answer, undermining the verifier's ability to distinguish between sound and flawed rationales.
We empirically show that in StrategyQA, only 19\% of LLM-generated solutions with correct answers have valid rationales.
Furthermore, we demonstrate that training a verifier on valid rationales significantly improves its ability to distinguish valid and flawed rationales.
To make a better verifier without extra human supervision, we introduce REPS (Rationale Enhancement through Pairwise Selection), a method for selecting valid rationales from candidates by iteratively applying pairwise self-evaluation using the same LLM that generates the solutions.
Verifiers trained on solutions selected by REPS outperform those trained using conventional training methods on three reasoning benchmarks (ARC-Challenge, DROP, and StrategyQA).
Our results suggest that training reliable verifiers requires ensuring the validity of rationales in addition to the correctness of the final answers, which would be critical for models assisting humans in solving complex reasoning tasks.

\end{abstract}

\begin{figure}[ht!]
\centering
\includegraphics[width=\columnwidth]{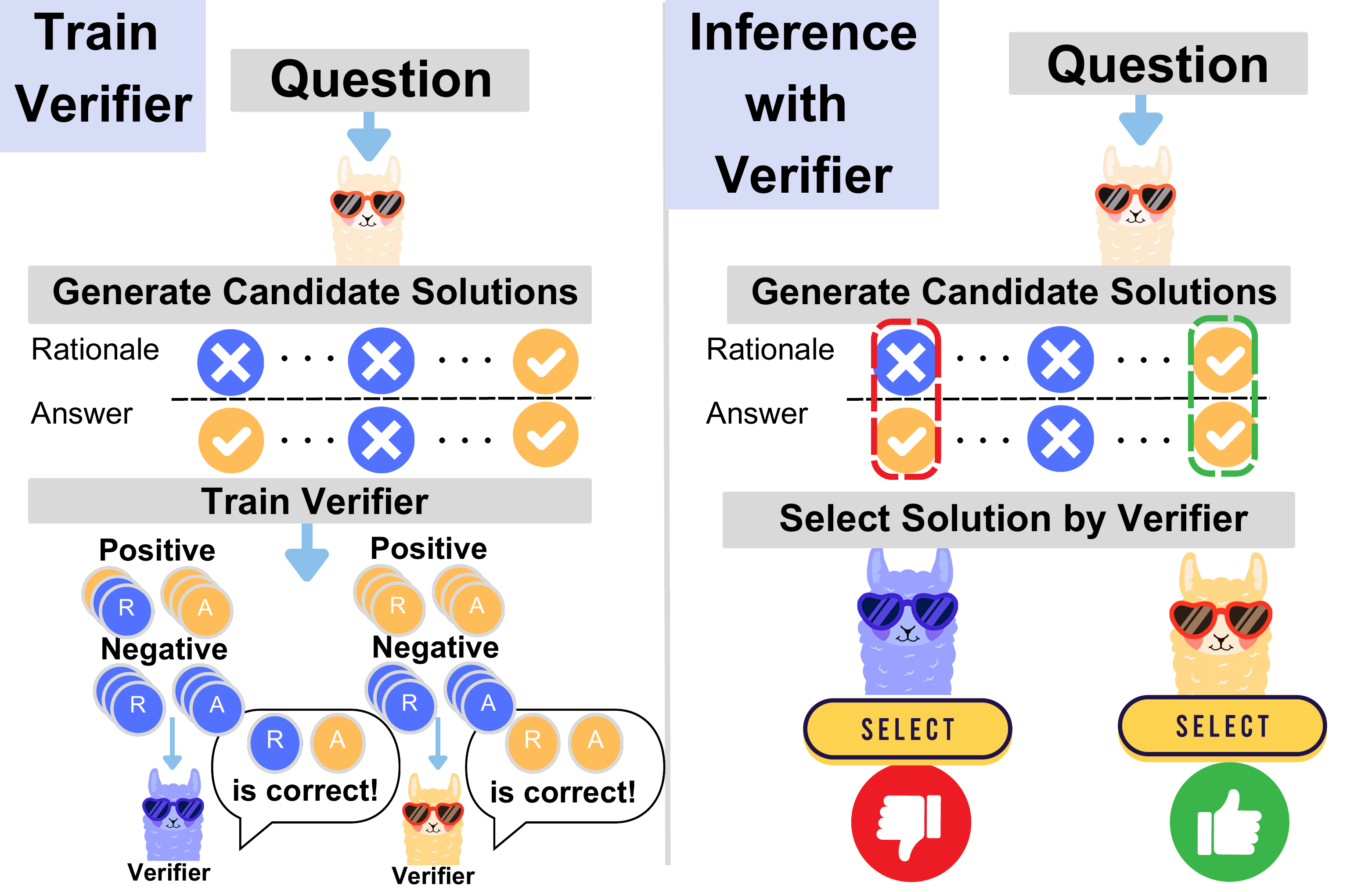}
\caption{
Importance of considering rationale quality in answer verification.
Verifiers trained on correct answers with flawed reasoning (blue) fail to identify valid rationales at inference. 
In contrast, verifiers trained on solutions with correct answers and rationales (yellow) can distinguish valid reasoning.
}
\label{fig:fig1}
\end{figure}

\section{Introduction}
Reasoning in large language models (LLMs) \cite{Brown-etal-2020-fewshot, chowdhery2022palm, openai2024gpt4} plays a vital role in their intelligent behavior \cite{wei2022emergent, kojima2022large}. 
Recent advancements, owing to progressive scaling \cite{kaplan2020scaling} and high-quality corpora \cite{penedo2024fineweb}, facilitate LLMs in solving complex reasoning tasks including mathematical reasoning and multi-hop reasoning. 
Further improvements are possible by prompting LLMs to generate intermediate reasoning steps that validate why the generated answer is correct \cite{wei2022chain}.
However, LLM-generated reasoning often exhibits factual and logical consistency error \cite{maynez-etal-2020-faithfulness, laban2023llms, lyu-etal-2023-faithful, jacovi2024chainofthought, dai2024imitation}.

Erroneous reasoning can be mitigated using a trained verifier model \cite{cobbe2021training} to select the best solution from multiple generated solutions, each consisting of an answer and its intermediate reasoning (rationale). 
This methodology delivers more accurate answers than a single-answer setting \cite{cobbe2021training, li-etal-2023-making, Hosseini2024-ph}.
However, existing studies mainly define the training labels for the verifier by solely checking the final answer. 
This approach may lead to a verifier that prefers faulty reasoning when LLMs generate the correct answers for wrong reasons (Figure~\ref{fig:fig1}). 
This is particularly problematic when models assist humans in complex tasks \cite{bowman2022measuring, saunders2022selfcritiquing}, wherein humans can not fully examine the model's feedback.

This raises two questions: (1) How often do models generate correct answers with flawed rationales? (2) Does training on flawed rationale as positive samples yield an untrustworthy verifier that cannot discriminate between flawed and valid rationale?

We investigate these questions using StrategyQA \cite{geva2021strategyqa} as the testbed (detailed in Section \ref{sec:3}). 
We use GPT-4 \cite{openai2024gpt4} to assess the validity of the rationales based on their factual and logical consistency with the annotated factual information. 
For the first question, we find that although 59\% of the model-generated solutions contain the correct answer, only 19\% of these correct-answer solutions are judged as having valid rationale, which is also supported by our manual inspection.
For the second question, we create training datasets with varying levels of reasoning quality for the verifier. 
The verifier's accuracy in selecting valid reasoning significantly increases as the quality improves.

To enhance the verifier's ability to discern the validity of rationales, we introduce REPS (Rationale Enhancement through Pairwise Selection), a training method that leverages LLMs' pairwise self-evaluation to select high-quality rationales from candidates. 
By refining the training data with these valid rationales, REPS enables to train more reliable and rationale-aware verifiers (Section~\ref{sec:4}).

Experiments across three reasoning datasets, ARC-Challenge \cite{clark2018think}, DROP \cite{dua-etal-2019-drop}, and StrategyQA demonstrate that verifiers trained on REPS consistently prefer solutions with valid rationales compared to those trained on datasets labeled solely based on the final answer's correctness. 
While REPS serves as an effective quality filter for rationales without extra human supervision, our analysis implies that excessive pairwise evaluation may amplify inherent preference biases in LLMs, such as favoring longer text (Sections~\ref{sec:5} and \ref{sec:6}). 
We hope future work will explore methods to mitigate these biases and further enhance the reliability of verifiers, in preparation for models becoming capable of solving tasks that humans cannot examine. 
This work takes the first step towards building trustworthy verifiers that can discern the validity of rationales.

Our contributions can be summarized as follows:

\begin{itemize}

\item We empirically demonstrate the crucial role of rationale quality in training reliable verifiers. 
Our analysis reveals that a large portion of correct answer solutions have invalid reasoning, and verifiers trained on such rationales tend to select solutions with similarly faulty reasoning. 

\item We introduce REPS, a method for selecting high-quality rationales by applying iterative pairwise self-evaluation.
Verifiers trained with REPS significantly outperform baseline verifiers in selecting solutions with valid rationale on ARC-Challenge, DROP, and StrategyQA.

\item Our analysis reveals that while iterative pairwise evaluation enhances the rationale quality, it can amplify inherent preference bias in LLM-based evaluators toward selecting longer rationales.\footnote{The code and data are available at \url{https://github.com/AkiraKawabata/REPS}.}

\end{itemize}

\section{Related Work}
\subsection{Answer Verification}
Recent studies have explored using verifier models for scoring or ranking generated solutions in domains such as mathematical reasoning \cite{cobbe2021training, Wang2023-zg, yu2023outcomesupervised, lightman2024lets, miao2024selfcheck}, coding \cite{ling2023deductive, ni2023lever}, and commonsense reasoning \cite{li-etal-2023-making, weng-etal-2023-large, zhang2024small}.
Verifiers are trained to distinguish between correct and incorrect solutions using several approaches, such as generative models \cite{pmlr-v202-korbak23a, asai2024selfrag}, reward models \cite{cobbe2021training}, or a combination \cite{rafailov2023direct, Hosseini2024-ph}.
When labeling solutions as ``correct'' and ``incorrect'', existing studies mainly rely on checking consistency with the final answer \cite{yu2023outcomesupervised, Hosseini2024-ph}.
A notable exception is \citet{li-etal-2023-making}, which attempts to extract reasonable reasoning steps from wrong answer solutions through comparisons of the steps with those in correct solutions.

However, reasoning paths generated by LLMs often contain logical or factual errors~\cite{lyu-etal-2023-faithful, turpin2023language, golovneva2023roscoe}, despite having a correct final answer \cite{jacovi2024chainofthought}.
This study challenges the assumption that a correct answer indicates a valid reasoning path to build rationale-aware verifiers.

\subsection{LLM-as-Judge}
Existing studies have explored using LLMs to evaluate natural language generation tasks \cite{chen-etal-2023-exploring-use, zheng2023judging, auto-correct-survey, kim2024prometheus, zeng2024evaluating}. 
These methods can be categorized into three types: providing natural language feedback \cite{madaan2023selfrefine, paul-etal-2024-refiner}, scoring with scalar values \cite{liu-etal-2023-g}, and comparing multiple outputs \cite{xie2023selfevaluation, qin2024large, liu2024aligning}.
LLM-based evaluation can also improve training by offering feedback as supervision \cite{xu-etal-2023-instructscore, liu-etal-2023-crystal, liu2024minds}, filtering the training dataset by score \cite{gulcehre2023reinforced} or using pairwise comparison \cite{bai2022constitutional}.
This study applies pairwise evaluation, which aligns better with human judgment than direct scoring \cite{wang2023large}, to refine the verifier's training data.

\section{Does Flawed Rationale Lead to Untrustworthy Verification?}
\label{sec:3}

LLMs may generate responses that contain correct answers but invalid reasoning paths \cite{jacovi2024chainofthought}.
Thus, labeling generated solutions as positive samples based on the correctness of the final answer, without considering the validity of the reasoning, may yield a verifier that fails to distinguish between sound and flawed reasoning.
Herein, we investigate this hypothesis by decomposing it into two questions:
(1) How often do LLMs generate correct answers with invalid reasoning? (Section \ref{sec:3-3})
(2) How does the rationale quality in the positive training data affect the verifier's performance in selecting valid reasoning? (Section \ref{sec:3-4})

\subsection{Task Setting of Answer Verification}
\label{sec:3-1}
In answer verification, a verifier model evaluates the correctness of multiple candidate solutions generated by an answer-generation model and selects the highest-scoring candidate as the final answer.
\label{sec:3-1}
\paragraph{Verification} 
A verifier model $M_v$ takes a solution $s$ generated by an answer-generation model $M_g$ for a question $q$, and returns the probability $p$ that the solution is correct. 
The solution $s$ consists of an answer $a$ and a reasoning path $r$ that represents the rationale justifying the answer $a$, i.e., $s = (a, r)$. 
The verifier's output probability can be expressed as:
$p = M_v(s \mid q)$.

\paragraph{Training Verifier Models} 
To train a verifier, we sample solutions from the generator $M_g$ for each question $q$ with temperature $T=0.7$.
These solutions are then classified as correct or incorrect based on arbitrary criteria (e.g., whether the answer matches the gold answer).
The verifier model $M_v$ is trained to judge the correctness of each solution $s$ as a reward model, following \citet{cobbe2021training}.
We use a binary cross-entropy loss function for training:
\begin{equation}
\small
\mathcal{L} = -\frac{1}{N}\sum_{i=1}^{N} [y_i \log(p_i) + (1-y_i) \log(1-p_i)],
\end{equation}
where $N$ is the number of training samples, $y_i$ is the binary label (1 for correct and 0 for incorrect) for the $i$-th solution, and $p_i$ is the probability predicted by the verifier for the $i$-th solution being correct.

\paragraph{Answer Selection by Verifiers}
At inference time, given a set of candidates ${s_1, s_2, \dots, s_n}$ generated by $M_g$ for a question $q$, the verifier model $M_v$ predicts the probability $p_i$ for each candidate $s_i$ and selects the highest:
\begin{equation}
s^* = \argmax_{s_i} M_v(s_i \mid q)
\end{equation}
where $s^*$ is the selected answer.

\subsection{Experimental Settings}
\label{sec:3-2}
\paragraph{Model} 
We use Llama-2 7B, a publicly available LLM, as the base model for both the answer generator and verifier owing to its popularity \cite{xie2023selfevaluation, Hosseini2024-ph}, thus facilitating a comparison with existing research.

\paragraph{Dataset} 
We use StrategyQA, a yes/no question dataset requiring multi-hop reasoning using factual knowledge. 
We select StrategyQA as our testbed because it requires broad factual knowledge and consistent reasoning, areas where LLMs often struggle \cite{jacovi2024chainofthought}. 
We split the official train (2,290 questions) subset into 1,603 training and 687 test questions, as the official test set does not contain gold answers.

\begin{figure*}[t!]
\centering
\begin{minipage}{0.48\textwidth}
    \centering
    \includegraphics[width=0.95\linewidth]{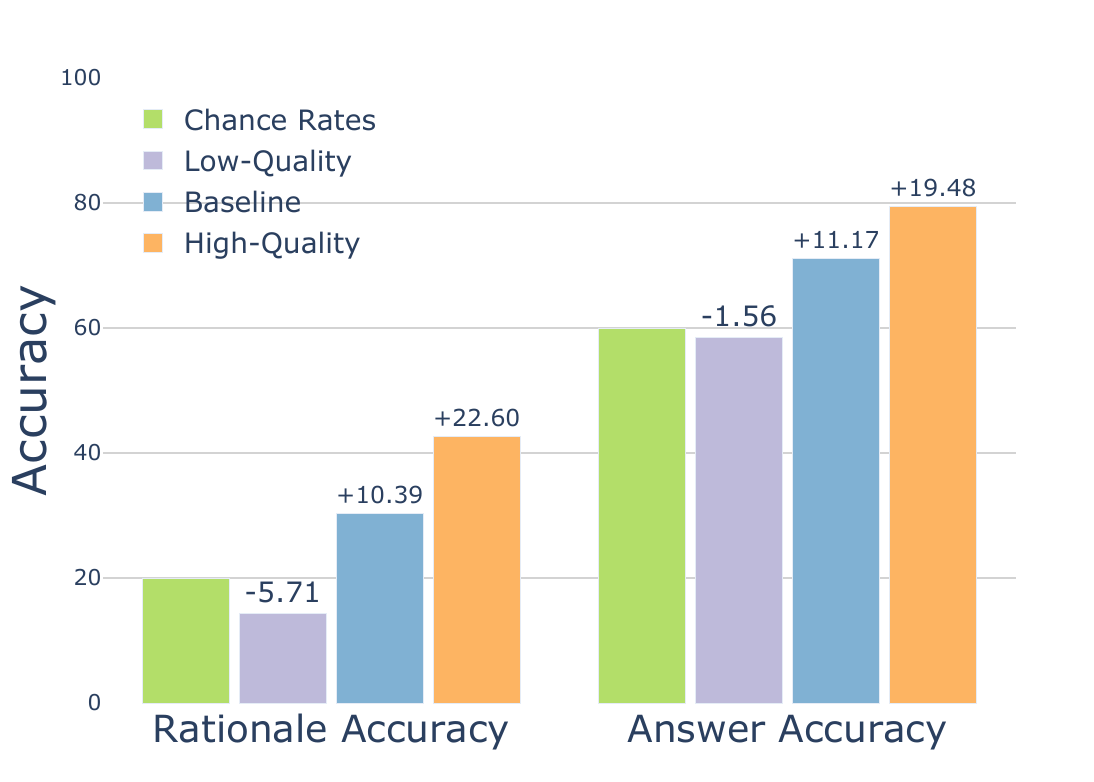}
    \caption{Rationale Accuracy (\%) and Answer Accuracy (\%) of verifier models trained on datasets with varying levels of rationale quality.}
    \label{fig:quality-ablation}
\end{minipage}\hfill
\begin{minipage}{0.48\textwidth}
    \centering
    \includegraphics[width=0.95\linewidth]{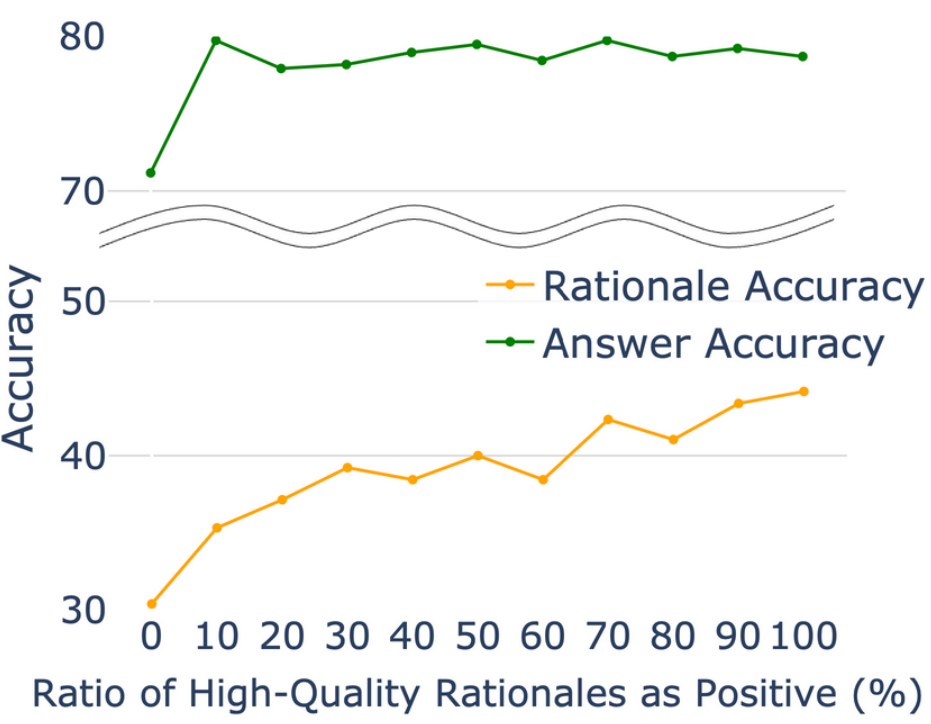}
    \caption{Rationale Accuracy (\%) and Answer Accuracy (\%) as a function of the ratio of high-quality rationales mixed into the baseline dataset.}
    \label{fig:validity-mixed}
\end{minipage}
\end{figure*}

\paragraph{Validity Evaluation}
We evaluate rationale quality based on factuality and logical consistency, following previous studies on model-generated reasoning evaluation \cite{ott2023thought, golovneva2023roscoe, radhakrishnan2023question, press-etal-2023-measuring, jacovi2024chainofthought}.
Factuality assesses the grounding of a model's reasoning in real-world knowledge, while logical consistency checks the coherence of the reasoning process.

We use GPT-4 to annotate the validity of generated rationales.
To ensure that GPT-4's judgment is not hallucinated, we provide it with supporting facts annotated in each question in StrategyQA.\footnote{Examples of GPT-4's annotations are provided in Appendix \ref{sec:appendix-a}.} 
GPT-4's validity annotations show a Cohen's kappa of 0.62 with our manual annotations on 100 randomly sampled rationales.\footnote{More details for our annotations are in Appendix \ref{sec:appendix-manual-annotation}.}

\subsection{How Often Does LLM Generate Correct Answers with Invalid Reasoning?}
\label{sec:3-3}
We examine how often model-generated solutions have invalid reasoning despite having correct answers.
For each question, we prompted the model with 6-shot examples to generate solutions $s$, consisting of a reasoning path $r$ and an answer $a$.

GPT-4's evaluation showed that while 59\% of the generated solutions contained the correct answer, only 19\% of those correct-answer solutions were judged as having valid reasoning paths.
This highlights the limitations of relying solely on answer correctness to determine the validity of a solution, as it does not guarantee sound reasoning.

\subsection{How Does Rationale Quality Affect Verifier Performance in Selecting Valid Reasoning?}
\label{sec:3-4}

We examine how rationale quality in positive samples affects verifier performance in selecting valid solutions. 
We conduct experiments with different levels of rationale quality.

\paragraph{Training Datasets}
We create three different training datasets for the verifier model.
Following previous studies \cite{cobbe2021training, li-etal-2023-making}, we consider solutions with incorrect answers as negative samples.

\begin{enumerate}
    \item \textbf{Low-quality setting}: Positive samples are created by replacing the final answer of incorrect solutions with the correct one, resulting in little to no valid reasoning.
    \item \textbf{Baseline setting}: Positive samples are solutions with correct final answers, regardless of reasoning quality.
    \item \textbf{High-quality setting}: Positive samples are solutions with correct answers and valid rationales, as validated by GPT-4.
\end{enumerate}

The quality of rationales in the positive samples is expected to increase from the Low-quality setting to the High-quality setting.
To collect valid solutions for the High-quality setting, we evaluate up to 20 generated solutions per question by GPT-4 in a zero-shot manner.
When any one solution passes the evaluation, we include the solution in the high-quality training dataset as a positive example. 
If none of the 20 solutions pass, the question is discarded. 
Consequently, we obtain 1,318 training examples.
We keep dataset sizes consistent across all settings for fair comparison.

\paragraph{Distribution-Controlled Test Set Design} 
We design a test set to evaluate the ability of the verifier to select solutions independently of the generator model's output distribution. 
For each question in the test set, we create a set of solution candidates consisting of:

\begin{itemize}
    \item One valid solution $s_{valid} = (r_{+}, a_{+})$ where $r_{+}$ denotes a valid rationale and $a_{+}$ is the corresponding correct answer.
    \item Two solutions $s_{correct} = (r_{-}, a_{+})$ with correct answer $a_{+}$ but invalid rationale $r_{-}$.
    \item Two solutions $s_{incorrect} = (r_{-}, a_{-})$ with incorrect answer $a_{-}$ and invalid rationale $r_{-}$.
\end{itemize}

\begin{figure*}[t!]
\centering
\includegraphics[width=\linewidth]{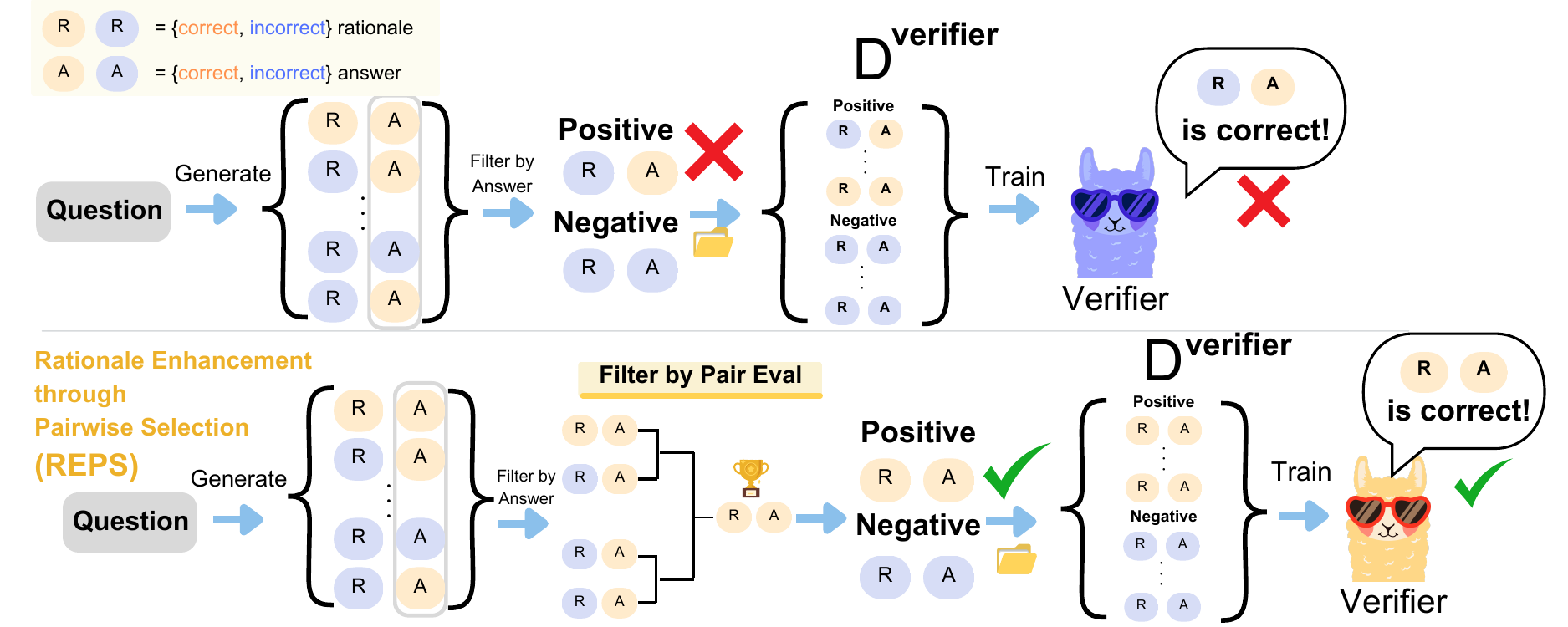}
\caption{
Rationale Enhancement through Pairwise Selection (REPS).
The generator model produces candidate solutions and filters out those with incorrect answers.
Unlike the conventional pipeline (top), REPS (bottom) employs a tournament-style pairwise evaluation to iteratively select the better solution. 
This refined solution is then used to train a rationale-aware verifier.
}
\label{fig:fig2}
\end{figure*}

$N_{\text{valid}}$, $N_{\text{correct}}$, and $N_{\text{incorrect}}$ denote the number of $s_{valid}$, $s_{correct}$, and $s_{incorrect}$, respectively.
We set $N_{\text{valid}} = 1$, $N_{\text{correct}} = 2$, and $N_{\text{incorrect}} = 2$. 
We report two metrics:

\begin{itemize}
\item \textbf{Rationale Accuracy (RA)}: The percentage of questions when the verifier selects the valid solution from the set of candidates.
\item \textbf{Answer Accuracy (AA)}: The percentage of questions when the verifier selects a solution with a correct answer (regardless of whether the rationale is valid or not) from the set of candidates.
\end{itemize}

Formally, these metrics can be expressed as:
\[
\text{RA} = \frac{1}{|D_{\text{test}}|} \sum_{i=1}^{|D_{\text{test}}|} \mathbb{I} \left[ \arg \max_{s \in S_i} M_v(s) = s_{\text{valid}} \right]
\]
\[
\text{AA} = \frac{1}{|D_{\text{test}}|} \sum_{i=1}^{|D_{\text{test}}|} \mathbb{I} \left[ \arg \max_{s \in S_i} M_v(s) \in s_{\text{good}} \right]
\]

where $s_{good} = s_{valid} \cup s_{correct}$, $D_{\text{test}}$ is the test dataset, $S_i$ is the set of answer candidates for the $i$-th question, and $\mathbb{I}[\cdot]$ is the indicator function.
The chance rates for RA and AA are 20\% and 60\%, respectively.

\subsection{Results}
\label{sec:3-5}
Figure \ref{fig:quality-ablation} shows RA and AA of verifier models trained on datasets with varying rationale quality. 
Training on the high-quality dataset, where the rationales are validated by GPT-4, results in a verifier that is significantly better at identifying valid reasoning compared to the baseline setting. 
Conversely, the low-quality dataset, where positive samples lack valid reasoning, leads to a verifier with RA near chance rates.

To further investigate the impact of rationale quality on the verifier's performance, we incrementally replace 10\% of the baseline dataset with high-quality rationales. 
Figure \ref{fig:validity-mixed} shows a clear trend: as the proportion of high-quality rationales increases, RA improves while AA remains largely stable.

These results highlight the importance of high-quality rationales in training verifier models, particularly in distinguishing between valid and flawed solutions.
Next, we propose a method for automatically constructing a training dataset with high-quality rationales.

\section{Rationale Enhancement through Pairwise Selection (REPS)}
\label{sec:4}
As shown in Section \ref{sec:3}, a large portion of LLM-generated answers contain flawed reasoning, even if the final answer is correct. 
Moreover, increasing the ratio of valid rationales in the positive samples improves the verifier’s ability to identify sound reasoning.
Thus, filtering out solutions with flawed rationales is crucial for building a reliable verifier.

We propose REPS, which uses LLMs' pairwise comparison to iteratively select high-quality rationales from diverse candidates.
The overall process of REPS is illustrated in Figure \ref{fig:fig2}.
For a question $q$, we first prompt the generator model $M_g$ to produce a set of candidate solutions $S = {s_1, s_2, \dots, }$, where each solution $s_i$ consists of an answer $a_i$ and a rationale $r_i$.
From this set, we select $N$ solutions with correct answers $a_{+}$ to form the candidate solution set for the tournament-style pairwise evaluation.

We use tournament-style pairwise evaluation to find the best rationale among the $N$ candidates.
In each round, the generator model $M_g$ acts as the evaluator that compares the rationales $(r_i, r_j)$ of two solutions $(s_i, s_j)$ $S$ times and selects the more factually grounded and logically consistent one as the winner through majority voting.
The evaluator is provided with the question, answer, two candidate rationales, and (depending on the dataset) passage and answer options.
Given hand-crafted few-shot exemplars, the evaluator model outputs a justification for which rationale is better and why, followed by the preferred rationale's index (1 or 2).\footnote{The prompt format used for the pairwise evaluation are provided in Appendix \ref{sec:appendix-reps-format}.}
To mitigate the position bias reported in LLM-based evaluations \cite{wang2023large}, we alternate the presentation order of rationale 1 and 2 across the $S$ comparisons within each round.

We perform this pairwise comparison for all $\frac{N}{2}$ pairs of solutions.
After the pairwise comparisons, the $\frac{N}{2}$ preferred rationales form a new set of candidate solutions $S'$, and the pairwise comparison process is repeated on this reduced set. 
This tournament-style elimination continues until only one rationale remains, which we consider the refined rationale $r^*$.
The refined rationale $r^*$, along with its corresponding answer $a^*$, is then added to the training set for the verifier model $M_v$ as a positive sample.
Repeating this for all questions refines the training dataset.

\begin{algorithm}[t!]
\caption{Rationale Enhancement through Pairwise Selection (REPS)}
\begin{algorithmic}
\REQUIRE Question $q$, Generator $M_{g}$, Num candidates $N$, Num voting $S$
\STATE $C \gets \{s_i \mid s_i \in M_g(q) \wedge s_i = (r_i, a_{+})\}$
\STATE $S \gets \text{sample}(C, N)$
\WHILE{$|S| > 1$}
\STATE $S' \gets \emptyset$
\FOR{$(s_i, s_j)$}
\STATE $s^* \gets \text{pairwise\_eval}(s_i, s_j)$
\STATE $S' \gets S' \cup \{s^*\}$
\ENDFOR
\STATE $S \gets S'$
\ENDWHILE
\STATE $s^* \gets S[0]$
\ENSURE $s^*$
\end{algorithmic}
\end{algorithm}

\begin{table*}[ht!]
\centering
\begin{tabular}{lcccccc}
\toprule
 & \multicolumn{2}{c}{ARC-Challenge} & \multicolumn{2}{c}{DROP} & \multicolumn{2}{c}{StrategyQA} \\
\cmidrule(lr){2-3} \cmidrule(lr){4-5} \cmidrule(lr){6-7}
       & Rationale Acc. & Task Perf. & Rationale Acc. & Task Perf. & Rationale Acc. & Task Perf. \\
\midrule
Baseline & 38.90 & 52.40 & 36.02 & 45.80 & 30.13 & 67.10 \\
REPS & \textbf{53.05} & \textbf{54.75} & \textbf{40.90} & \textbf{46.90} & \textbf{38.96} & \textbf{67.25} \\
\bottomrule
\end{tabular}
\caption{
Rationale Accuracy (\%) and Task Performance (\%) of the baseline verifier and REPS.
REPS consistently outperforms the baseline in selecting valid rationales while maintaining or slightly improving Task Performance.
}
\label{tab:main-result}
\end{table*}

\section{Experiment}
\label{sec:5}

\begin{figure*}[ht!]
\centering
\includegraphics[width=0.45\linewidth]{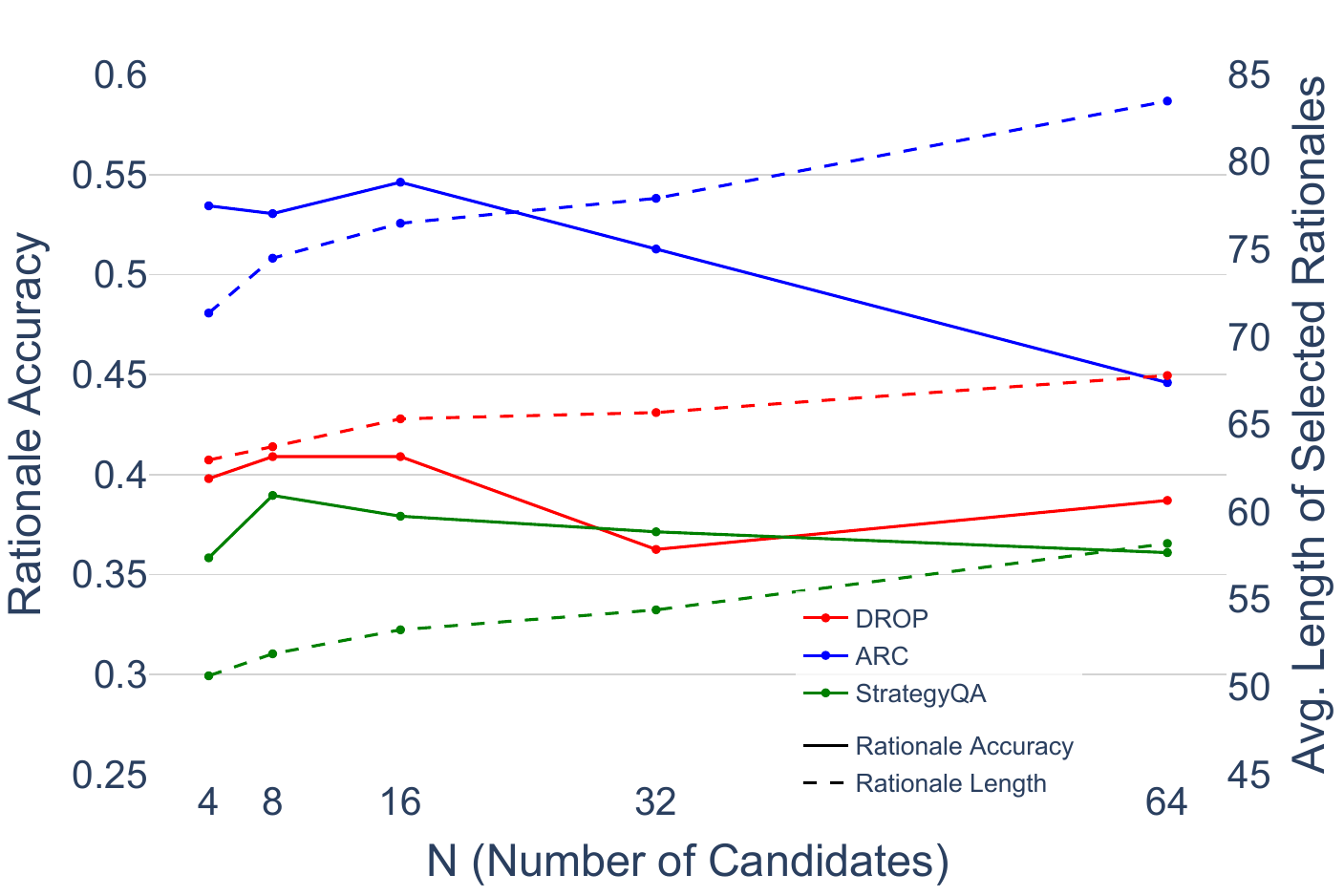}
\hfill
\includegraphics[width=0.45\linewidth]{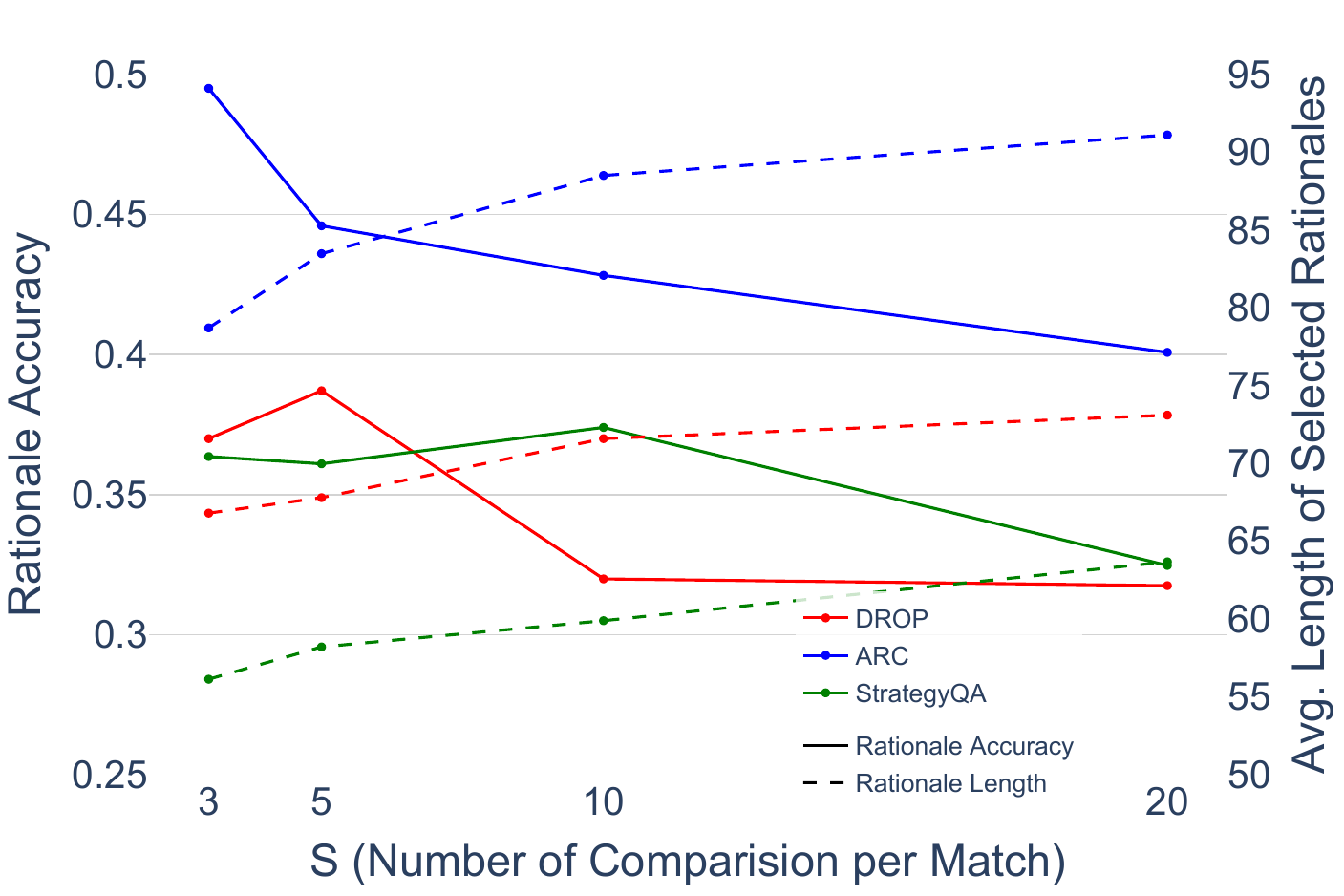}
\caption{The effect of varying the number of candidate solutions ($N$) and the number of pairwise comparisons per match ($S$) on the Rationale Accuracy (\%) and average length of selected rationales. 
Increasing N and S leads to a decrease in Rationale Accuracy and an increase in the average length of selected rationales.}
\label{fig:N_and_S}
\end{figure*}

\subsection{Setup}

\paragraph{Model}
We employ Llama-2 7B as the backbone for both the generator ($M_g$) and verifier ($M_v$) models. 
We report the performance of two verifiers: the baseline verifier and the REPS verifier. 
The baseline verifier is trained using the conventional approach, labeling solutions as positive based solely on the correctness of the final answer. 
The REPS verifier is trained on a dataset where solutions refined through REPS are labeled as positive. \footnote{Details of the training and the inference are provided in Appendix \ref{sec:appendix-training} and \ref{sec:appendix-inference}, respectively.}

\paragraph{Metrics}
We report two metrics: Rationale Accuracy and Task Performance. 
Rationale Accuracy is evaluated on a distribution-controlled test set (Section~\ref{sec:3-3}).
The test set is constructed by sampling $N_\text{valid}=1$, $N_\text{correct}=2$, and $N_\text{incorrect}=2$ solutions for each question. 
Valid solutions are selected by GPT-4, and questions for which none of the 20 evaluated solutions are judged valid are discarded as done in Section \ref{sec:3}. 
Task Performance is evaluated by sampling five solutions for each question, ranking them with the verifier, and selecting the answer in the highest-scoring solution.

\paragraph{Datasets}
We evaluate our method on three diverse datasets that assess different aspects of reasoning capabilities:

\begin{enumerate}
\item StrategyQA: Dataset consisting of questions that require multi-hop reasoning using factual knowledge to arrive at the correct answer. 
We use 1,603 data points for training, 687 for testing Task Performance, and 385 of the test questions to evaluate Rationale Accuracy.
We report accuracy as Task Performance.

\item ARC-Challenge: Challenging subset of the AI2 Reasoning Challenge (ARC) dataset that evaluates commonsense reasoning about scientific knowledge. 
We use 1,119 questions for training, 1,172 for testing Task Performance, and 509 of the test questions to evaluate Rationale Accuracy.
We report accuracy as Task Performance.

\item DROP: Reading comprehension dataset that requires arithmetic reasoning to answer the questions. 
We select 2,000 questions for training, 1,000 for testing Task Performance, and 819 for evaluating Rationale Accuracy.
We report Exact Match as Task Performance.
\end{enumerate}

\paragraph{Additional References for GPT-4's Annotation}
We provided additional references for each dataset to support GPT-4's annotation of valid solutions. 
For StrategyQA, as described in Section \ref{sec:3}, we provide the supporting facts annotated in the dataset. 
For ARC, we provided the top five Wikipedia paragraphs with the highest BM25 scores against the concatenated question and answer options. 
We did not provide any additional references for DROP, as it is a reading comprehension task that does not require external factual knowledge.

\paragraph{REPS Parameters}
We set the number of pairwise comparisons per round ($S$) to 5 and the number of candidate solutions ($N$) to 8 for ensuring diversity and robustness in voting.

\section{Results and Analysis}
\label{sec:6}

\subsection{Effectiveness of REPS}
\paragraph{Main Results}
Table \ref{tab:main-result} shows REPS improves Rationale Accuracy without affecting Task Performance across all datasets.
The improvement is particularly significant in ARC and StrategyQA, where the REPS-trained verifier substantially outperforms the baseline, with improvements of 14.1\% and 8.8\%, respectively. 
This improvement can be attributed to the nature of the reasoning required in these datasets. 
In arithmetic reasoning tasks like DROP, which involve more deductive reasoning, it is rare for the model to arrive at the correct answer using flawed reasoning.
Thus, the correctness of the final answer can be a good indicator of the validity of the reasoning process. 
In contrast, inductive and abductive reasoning tasks like ARC and StrategyQA allow more room for flawed reasoning to reach correct answers, emphasizing the importance of evaluating the intermediate reasoning.

\begin{figure}[t!]
\centering
\includegraphics[width=1.05\linewidth]{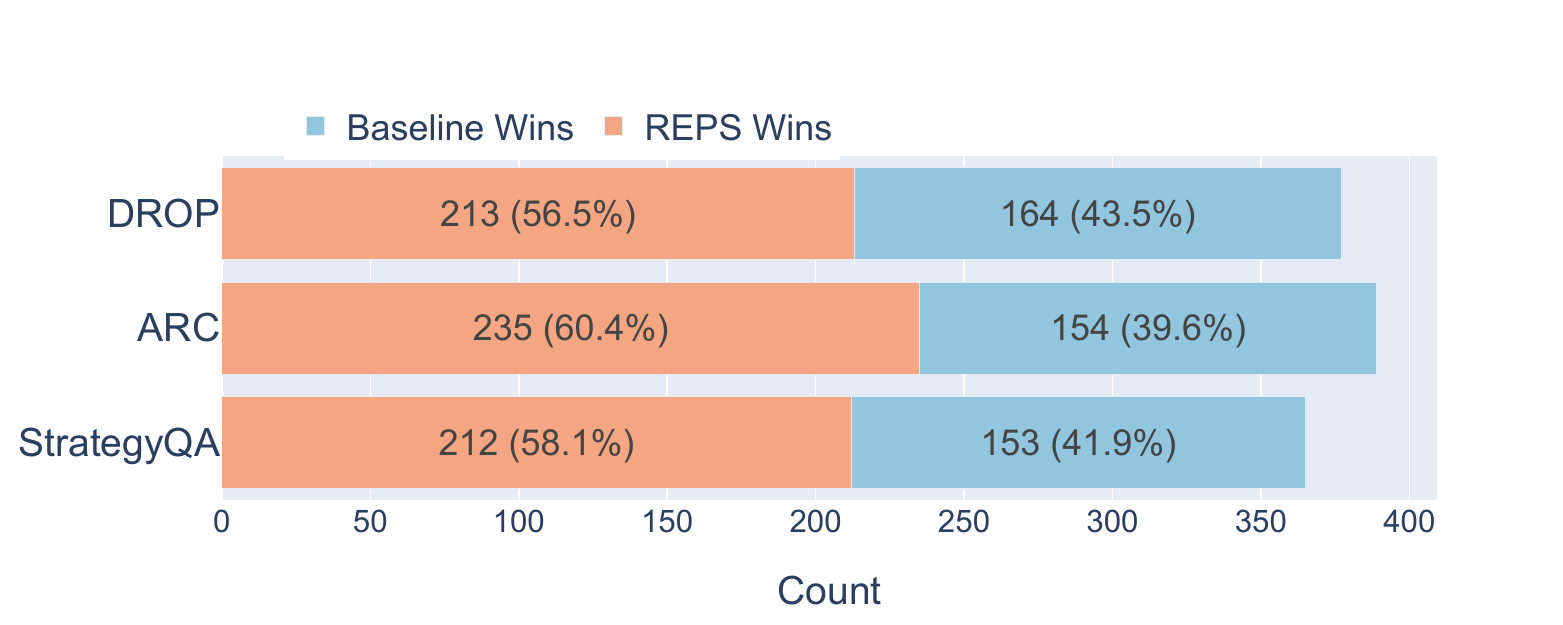}
\caption{Win rate of REPS vs. baseline verifier when their selected rationales are compared head-to-head using GPT-4 as the judge.}
\label{fig:win-rate}
\end{figure}

\paragraph{Win Rate by GPT-4}
We conduct a head-to-head evaluation using GPT-4 to compare the rationales selected by REPS and the baseline. 
For each question, we generate five candidate solutions, from which REPS and the baseline each select one. 
GPT-4 compares rationales for selected solutions, excluding those where both methods choose the same. 
In each pairwise evaluation, GPT-4 chooses the more factually grounded and logically consistent rationale, given the question and answer. 
We alternate the presentation order of the rationales selected by REPS and the baseline to mitigate position bias. 
Figure \ref{fig:win-rate} shows REPS-selected rationales consistently outperform baseline-selected across datasets, with 56-60\% win rates. 
Manual annotation of 100 random rationale pairs also confirms this trend, with REPS win rates of 63\%, 60\%, and 61\% for ARC, DROP, and StrategyQA, respectively.\footnote{Appendix \ref{sec:appendix-manual-annotation} shows details in our manual annotation.}
REPS chooses better rationales than the baseline, even without guaranteed valid rationales among candidates.

\subsection{Effects of Parameter Choice}
We study how parameters in REPS, candidate pool size ($N$), and pairwise comparisons per round ($S$) affect Rationale Accuracy.
A larger $N$ is expected to increase the diversity of the candidate pool. 
Similarly, a higher $S$ should enhance the reliability of the majority voting process. 
Thus, increasing both factors are expected to contribute to a higher Rationale Accuracy. 
Figure \ref{fig:N_and_S} illustrates the Rationale Accuracy as $N$ varies from 4 to 64 and $S$ from 3 to 20. 
Unexpectedly, Rationale Accuracy decreases as $N$ and $S$ grow.
This phenomenon can be seen as the emergence of biases toward the superficial cue, i.e., the length of rationale, in model-based evaluation, as shown in Figure \ref{fig:N_and_S}.
As $N$ and $S$ increase, i.e., the number of pairwise evaluations per tournament grows, the length of the selected reasoning paths also tends to increase. 
This implies model-based evaluation's bias for longer answers \cite{zheng2023judging, koo2023benchmarking, dubois2024lengthcontrolled} amplifies over repeated pairwise evaluations, causing deviation from valid rationale distribution.

One approach to handle the bias amplification is to treat the parameters $N$ and $S$ in REPS as hyperparameters that can be tuned using a validation set. 
Adjusting these parameters makes it possible to identify an optimal balance between effective rationale filtering and the minimization of bias amplification.

\begin{table}[t!]
\centering
\begin{tabular}{lccc}
\toprule
 & ARC & DROP & StrategyQA \\
\midrule
G-EVAL & 44.99 & 40.66 & 30.39 \\
REPS & \textbf{53.05} & \textbf{40.90} & \textbf{38.96} \\
\bottomrule
\end{tabular}
\caption{Rationale Accuracy (\%) of verifiers trained on rationales selected by REPS vs. G-EVAL.}
\label{tab:ablation-pairwise}
\end{table}

\subsection{Importance of Pairwise Evaluation} 
We investigate whether the iterative pairwise evaluation employed by REPS produces more effective training samples for rationale-aware verifiers compared to single-answer evaluation.
We use \textsc{G-EVAL} \cite{liu-etal-2023-g} as a representative of the single-answer evaluation, which assigns a weighted score to each solution $s$ based on the probabilities of outputting score tokens:

\begin{equation}
\text{g}(s) = \sum_{t_i=1}^{5} p(t_i \mid s) \times t_i
\end{equation}

where $t_i$ represents the i-th score token (e.g., ``1’’), and $p(t_i \mid s)$ is the probability of outputting that token given the solution $s$.
For each of the $N$ candidate solutions, we compute the score $S$ times and select the reasoning path with the highest average score.
For fair comparison, we set candidate solutions $N$ to 8 and evaluations per solution $S$ to 5 for both methods.
Table \ref{tab:ablation-pairwise} shows that verifiers trained on REPS-selected rationales outperform those trained on \textsc{G-EVAL}-selected ones in identifying valid reasoning.
This result demonstrates that REPS’s iterative pairwise comparison is more effective than direct scoring in providing higher-quality training samples for verifiers.

\subsection{Error Types in Invalid Reasoning Paths}
\begin{table}[ht]
\centering
\begin{tabular}{lccc}
\toprule
Error Type & ARC & StrategyQA & DROP \\
\midrule
Factual only & 16 & 15 & 35 \\
Logical only & 43 & 46 & 22 \\
Fact. and Logic. & 32 & 31 & 24 \\
Other & 2 & 5 & 9 \\
No errors & 7 & 3 & 10 \\
\bottomrule
\end{tabular}
\caption{Distribution of error types (\%) in reasoning paths judged as invalid by GPT-4.}
\label{tab:error_distribution}
\end{table}
To gain insight into the types of errors present in reasoning paths judged as invalid by GPT-4, we conduct a manual analysis of 100 randomly sampled invalid reasoning paths for each dataset (ARC, StrategyQA, and DROP). 
The errors are categorized into four types: (1) both factual and logical errors, (2) factual errors only, (3) logical errors only, and (4) other errors that are neither factual nor logical (e.g., incomplete explanations or mere repetitions of the question). 
We define factual errors as reasoning that uses information contradicting the supporting facts or passage provided for the question. 
Logical errors are defined as reasoning that is irrelevant to a valid explanation or draws conclusions that cannot be derived from the preceding steps. 
Table \ref{tab:error_distribution} presents the distribution of these error types across the three datasets.

In DROP, factual errors are notably more prevalent (35\% factual only and 24\% both factual and logical). 
The model often struggles with extracting necessary information from the given passage, frequently using hallucinated or incorrect information in its reasoning process.

Conversely, ARC and StrategyQA exhibit a higher proportion of logical errors (43\% and 46\% logical only, respectively). 
The model often fails to identify the relevant information necessary to solve the problem, or misunderstands what can be inferred from commonsense knowledge.

\section{Discussion}
\paragraph{The Discrepancy Between Rationale Accuracy and Task Performance}

While REPS effectively improves Rationale Accuracy of verifiers, it does not significantly enhance overall Task Performance. 
This aligns with Section \ref{sec:3-3} and recent findings suggesting a weak link between rationale quality and answer correctness \cite{wiegreffe-etal-2022-reframing, jacovi2024chainofthought}.
This unfaithfulness becomes particularly critical in the context of scalable oversight \cite{bowman2022measuring}, where humans may find it challenging to evaluate the correctness of the model's outputs. 
In such scenarios, humans might be misled by seemingly plausible yet unfaithful rationales. 
Future research should focus on improving rationales regarding quality and faithfulness.

\section{Conclusion}

We investigate how rationale quality affects a verifier's ability to select valid answers.
We empirically demonstrate that many model-generated rationales contain errors, even when the final answer is correct, leading to untrustworthy answer verification. 
We introduce REPS, a method that uses LLMs' pairwise comparison to iteratively refine generated rationales.
Experiments on three reasoning datasets show that REPS significantly outperforms baseline verifier models, particularly in selecting solutions with valid rationales. 
Our analysis shows iterative pairwise evaluation improves rationale quality but may amplify LLM-based evaluators' biases.

\section*{Limitations}

\paragraph{Dataset Diversity} 
The experiments in this study are limited to three reasoning datasets: ARC-Challenge, DROP, and StrategyQA. 
While these datasets cover various aspects of reasoning, they may not be representative of all reasoning tasks. 
To further validate the effectiveness of REPS, it would be beneficial to evaluate the method on a more diverse set of datasets, such as those involving coding \cite{chen2021evaluating} or instruction following \cite{zheng2023judging}.

\paragraph{Reliance on GPT-4 for Judging}
Although we confirmed a high agreement between GPT-4 and human judgments, GPT-4 may still be subject to biases not present in human evaluations, potentially leading to invalid evaluation.

\paragraph{Limited Training Data}
We use a relatively small amount of training data, ranging from 1,000 to 2,000 instances per dataset. 
It would be valuable to investigate REPS's scalability on larger datasets.

\paragraph{Limited Model Size}
This study focuses on a single model, Llama-2 7B, as a case study for the generator and verifier models. 
It would be valuable to explore the impact of model size and architecture on REPS, as larger models may provide more accurate and coherent pairwise evaluations, leading to further improvements in the selected rationales.

\section*{Acknowledgments}
We would like to thank the anonymous reviewers for their constructive feedback and Hideaki Tamori for his valuable comments.
This work was supported by JSPS KAKENHI Grant Number JP22K17954.

\bibliography{anthology, custom}

\appendix

\section{Training Details}
\label{sec:appendix-training}
We train the verifier models using Llama-2 7B as the base model. The hyperparameters for training are as follows: batch size per device of 32, 10 training epochs, learning rate of $3 \times 10^{-7}$, AdamW optimizer, and a random seed of 42. The models are trained on three NVIDIA A100 GPUs using mixed precision (bfloat16) to reduce memory usage and training time.

\section{Inference Details}
\label{sec:appendix-inference}

\subsection{Answer Generation}
\label{sec:appendix-inference-answer-generation}
For generating candidate answers, we use the Llama-2 7B model with a temperature of 0.7. The number of few-shot examples provided to the model varies depending on the dataset. For ARC, we use 4 examples; for DROP, we use 4 examples; and for StrategyQA, we use 6 examples. The generated answers are then fed into the verifier models for scoring and selection.

\subsection{REPS Pairwise Evaluation}
\label{sec:appendix-inference-reps-pairwise}
In REPS, we perform pairwise comparisons of the generated rationales using the Llama-2 7B model with a temperature of 0.7. The number of few-shot examples used for each dataset is determined by the maximum number of tokens that can fit within the model's context size of 4096. For ARC, we use 8 examples; for DROP, we use 3 examples; and for StrategyQA, we use 7 examples. The pairwise comparisons are conducted iteratively, with the winning rationales from each round advancing to the next round until a single rationale remains.

\subsection{GPT-4 Evaluations}
\label{sec:appendix-inference-gpt4-eval}

For evaluating the validity of the generated rationales and measuring the win rates between the REPS-selected rationales and the baseline, we employ the GPT-4 model with a temperature of 0.0. We use the \texttt{gpt-4-0314} version of OpenAI API.
The instructions used for evaluating the rationales are following:

\begin{quote}
You are a helpful assistant who evaluates explanations’ factual and logical consistency. Given a question and the corresponding reference information (evidence), you will assess whether the explanation meets the following criteria with justifications:

(1) Factual correctness: The explanation should be factually correct based on the given evidence. If the explanation contains claims not directly supported by or contradicting the evidence, it should be judged incorrect. The explanation should not include information or make assertions not mentioned in the evidence.

(2) Logical coherence: The explanation should follow a clear and logical chain of reasoning. Each step should be appropriately justified based on the evidence and preceding steps. Leaps in logic, assumptions, or opinions not grounded in the given information are unacceptable.

If the explanation meets all of the above criteria, output True. Otherwise, the output will be false if the explanation fails to meet any criteria. Your output must be only either True or False. Aim to prefer False if there are doubts about whether the criteria are fully satisfied.
\end{quote}

\section{REPS Pairwise Evaluation Prompts}
\label{sec:appendix-reps-format}

We use dataset-specific formats for pairwise evaluation prompts in REPS, with a common instruction across all datasets:
\begin{quote}
You are a helpful assistant who evaluates explanations' factual accuracy and logical consistency.
Given a question and an answer, decide which of the two provided explanations is more factually grounded and logically valid.
Your output must be 1 or 2, where 1 corresponds to the first explanation, and 2 corresponds to the second explanation.
\end{quote}

Pairwise evaluation is performed by providing the model with prompts in the following format:
\begin{quote}
\{\{instruction\}\}
\\
Question: \{\{question\}\}
\\
Answer: \{\{gold answer\}\}
\\
Explanation 1: \{\{explanation1\}\}
\\
Explanation 2: \{\{explanation2\}\}
\\
Justification: \{\{justification\}\}
\\
Preferred Explanation: \{\{1 or 2\}\}
\\
\end{quote}

\section{Examples of GPT-4's Validity Judge}
\label{sec:appendix-a}

We provide examples of GPT-4's judgments on the validity of reasoning paths generated by the Llama-2 7B in Table \ref{tab:example_gpt4judge}.
\textcolor{red}{Logical errors} in the reasoning paths are highlighted in red, while \textcolor{blue}{factual errors} are highlighted in blue.

\begin{table*}[t!]
\centering
\begin{tabular}{p{0.95\linewidth}}
\toprule
\textbf{ARC Example:} \\
\textbf{Question:} \\
Farmers in Wyoming were concerned because some of their chickens were being preyed upon by hawks that lived in areas around their ranches. The farmers grouped together and hunted the hawks until they were no longer in their area. Which would most likely happen next? \\
\textbf{Choices:} \\
A. The chicken population would go down. B. Populations of mice and rats would increase. C. Another bird of prey would replace the hawk. D. The chickens would have a lower rate of disease. \\
\textbf{Valid Rationale:} \\
Removing the hawks from the area could affect the populations of other animals.
If hawks are not present in an area, they are not eating mice and rats.
Without hawks, more mice and rats would survive.
This could increase the populations of mice and rats in surrounding areas.
So, the correct answer is, "B. Populations of mice and rats would increase." \\
\textbf{Invalid Rationale:} \\
When the hawks were no longer in the area, the chickens had to \textcolor{blue}{compete with other predators for food}. This led to a \textcolor{red}{decrease in the chicken population}.
Since the hawks were no longer around, the populations of mice and rats would increase.
\textcolor{red}{Therefore, the chickens would have a lower rate of disease} because there were fewer of them to catch diseases.
Therefore, the answer is B.\\
\midrule
\textbf{DROP Example:} \\
\textbf{Passage:}\\
Skopje, as the Republic of Macedonia as a whole, is characterised by a large ethnic diversity. The city is located in a region where Ethnic Albanians and Macedonians meet, and it welcomed Romani, Turks, Jews and Serbs throughout its history. Skopje was mainly a Muslim city until the 19th century, when large numbers of Christians started to settle there. According to the 2002 census, Macedonians were the largest ethnic group in Skopje, with 338,358 inhabitants, or 66.75\% of the population. Then came Albanians with 103,891 inhabitants (20.49\%), Roma people with 23,475 (4.63\%), Serbs (14,298 inhabitants), Turks (8,595), Bosniaks (7,585) and Vlachs (2,557). 8,167 people did not belong to any of these groups.\\
\textbf{Question:} \\
Who was the second most populous group in Skopje? \\
\textbf{Valid Rationale:} \\
The passage states that Albanians were the second most populous group in Skopje, with 103,891 inhabitants.The passage then goes on to list other groups, and mentions that the largest group is Macedonians.This indicates that the second most populous group is Albanians.\\
\textbf{Invalid Rationale:} \\
The passage mentions that Macedonians were the largest ethnic group in Skopje, with 338,358 inhabitants, or 66.75\% of the population. According to the passage, the second most populous group in Skopje was Albanians, with 103,891 inhabitants, or 20.49\% of the population. Albanians were the second most populous group, and the largest ethnic group was Macedonians, and \textcolor{red}{therefore, the Macedonians were the second most populous group in Skopje.} \\
\midrule
\textbf{StrategyQA Example:} \\
\textbf{Question:} \\
Do members of the Supreme Court of the United States have longer terms than most senators? \\
\textbf{Valid Rationale:} \\
A member of the Supreme Court of the United States is appointed to serve for life. A senator's term is six years. Thus, a member of the Supreme Court of the United States has a longer term than most senators. So the answer is yes. \\
\textbf{Invalid Rationale:} \\
The terms of Supreme Court justices are \textcolor{blue}{9 months}, which is \textcolor{red}{longer than the 6-year} term of a senator. Therefore, the answer is yes. \\
\bottomrule
\end{tabular}
\caption{Examples of GPT-4's validity judgments on reasoning paths generated by Llama-2 7B across three datasets: ARC (top), DROP (middle), and StrategyQA (bottom). The valid rationales demonstrate sound reasoning, while the invalid rationales contain \textcolor{blue}{factual errors} and \textcolor{red}{logical inconsistencies}.}
\label{tab:example_gpt4judge}
\end{table*}

\begin{table*}[t!]
\centering
\begin{tabular}{p{0.95\linewidth}}
\toprule
\textbf{Question:} \\
Did University of Pittsburgh founder have great deal in common with Judith Sheindlin? \\
\textbf{Rationale selected by REPS:} \\
Judith Sheindlin is 76 years old. The founder of the University of Pittsburgh died in 1811, which was 128 years before Judith Sheindlin was born. Thus, the founder of the University of Pittsburgh did not have a great deal in common with Judith Sheindlin. \\
\textbf{Rationale selected by Baseline:} \\
The founder of the University of Pittsburgh was William Pitt. Judith Sheindlin ("The Judge") has a first name that is similar to William. Thus, they have a great deal in common. \\
\midrule
\textbf{Question:} \\
Is Disney associated with Los Angeles County? \\
\textbf{Rationale selected by REPS:} \\
Disney has theme parks in Los Angeles County, California. Thus, Disney is associated with Los Angeles County. \\
\textbf{Rationale selected by Baseline:} \\
Disney is a large company that has many locations. Disneyland is located in Anaheim, which is 24 miles south of Los Angeles. Thus, Disney does not have a location in Los Angeles County. \\
\midrule
\textbf{Question:} \\
Could a bee hummingbird balance a scale with a single pea on it? \\
\textbf{Rationale selected by REPS:} \\
A bee hummingbird weighs about 0.4g. A single pea weighs about 0.2g. Thus, a bee hummingbird could balance a scale with a single pea on it. \\
\textbf{Rationale selected by Baseline:} \\
The mass of a bee is about 1.8 milligrams. The mass of a pea is about 1.5 grams. Thus, a bee could not balance a scale with a single pea on it. \\
\midrule
\textbf{Question:} \\
Can photography be considered abstract art? \\
\textbf{Rationale selected by REPS:} \\
Photography is art, but it is not abstract art. Photography can be realistic or non-realistic. Abstract art is non-realistic. Thus, photography cannot be considered abstract art. \\
\textbf{Rationale selected by Baseline:} \\
Photography is a type of art. Abstract art is a type of art. Thus, photography can be considered abstract art. \\
\midrule
\textbf{Question:} \\
Are aggressive bumblebees suicidal? \\
\textbf{Rationale selected by REPS:} \\
Bumblebees can become aggressive when they are protecting their colony. These bees can be very aggressive. Thus, aggressive bumblebees are not suicidal. \\
\textbf{Rationale selected by Baseline:} \\
Bumblebees are both aggressive and suicidal. \\
\bottomrule
\end{tabular}
\caption{Examples of rationales selected by REPS and the baseline verifier for questions from StrategyQA.}
\label{tab:example_strategyqa_reps_baseline}
\end{table*}

\begin{table*}[t!]
\centering
\begin{tabular}{p{0.95\linewidth}}
\toprule
\textbf{Question:} \\
Devil facial tumor disease (DFTD) is a disease that is decimating the population of Tasmanian devils. The disease passes from one animal to another through bites and is caused by parasites. The parasites cause cancerous tumors that spread throughout an infected animal's body and kill it. What is the best description of DFTD? \\
\textbf{Choices:} \\
A. a non-infectious, cell-cycle disease B. an infectious, cell-cycle disease C. a non-infectious, chronic disease D. an infectious, chronic disease \\
\textbf{Rationale selected by REPS:} \\
DFTD is a deadly infectious disease that is spread by parasites and causes cancerous tumors that spread throughout an infected animal's body. Thus, DFTD is an infectious, cell-cycle disease. \\
\textbf{Rationale selected by Baseline:} \\
The disease is caused by parasites, which are foreign organisms that invade a host body and cause disease. The parasites cause cancerous tumors that spread throughout the body and kill the host. Therefore, DFTD is an infectious, cell-cycle disease. \\
\midrule
\textbf{Question:} \\
A type of small mammal from the mountain regions of the western United States makes its home out of piles of rock. During summer months, the mammal places grasses and seeds in protected places in the rock piles. Which of the following is the most likely reason for this behavior? \\
\textbf{Choices:} \\
A. to prepare for migration before winter B. to provide warmth during the cold winter months C. to store food that will be eaten over the winter months D. to protect the grasses and seeds from decay before winter \\
\textbf{Rationale selected by REPS:} \\
This is the behavior of a prairie dog, which stores grass, seeds, and other food items in rock piles during summer months. Prairie dogs will consume the stored food during winter months. Thus, the prairie dog's behavior is to store food for use during the winter months. \\
\textbf{Rationale selected by Baseline:} \\
To protect and preserve its food, the mammal buries grasses and seeds in piles of rock. Grasses and seeds decay over time. Thus, the mammal buries its food to protect it from decay. \\
\midrule
\textbf{Question:} \\
According to cell classification, prokaryotic cells are separated from eukaryotic cells. Which feature is often used to distinguish prokaryotic cells from eukaryotic cells? \\
\textbf{Choices:} \\
A. life processes B. size differences C. plasma membranes D. energy molecules \\
\textbf{Rationale selected by REPS:} \\
Prokaryotes are bacteria and archaea, whereas eukaryotes are animals, plants, fungi, and other multicellular organisms. Prokaryotes do not contain membrane-bound organelles (structures) such as the nucleus. Therefore, the presence or absence of membrane-bound organelles is an important distinguishing feature of prokaryotes and eukaryotes. \\
\textbf{Rationale selected by Baseline:} \\
Prokaryotic cells lack a nucleus, while eukaryotic cells contain a nucleus. Thus, prokaryotic cells are distinguished from eukaryotic cells by the presence of a nucleus. \\
\bottomrule
\end{tabular}
\caption{Examples of rationales selected by REPS and the baseline verifier for questions from ARC-Challenge.}
\label{tab:example_arc_reps_baseline}
\end{table*}

\begin{table*}[t!]
\centering
\begin{tabular}{p{0.95\linewidth}}
\toprule
\textbf{Passage:} \\
Taunton has four art galleries: Taunton Art Association , Hughes/Donahue Gallery, Art Euphoric, and the Trescott Street Gallery. The Taunton Art Association founded in 1973, but had it roots at the Girl's Club in the early 1960s. Hughes/Donahue Gallery founded in 2007, a local community gallery serving local Taunton artists, surrounding areas of Southeastern Massachusetts and including the cities of Providence, and Washington DC. Art Euphoric founded in 2008 has both visual and craft exhibits and sales. The Trescott Street Gallery founded in 2012, primarily a visual arts gallery, but also exhibits crafts. \\
\textbf{Question:} \\
Which gallery was founded first, Hughes/Donahue Gallery or Art Euphoric? \\
\textbf{Rationale selected by REPS:} \\
The passage mentions that Hughes/Donahue Gallery was founded in 2007, and Art Euphoric was founded in 2008. The gallery that was founded first is Hughes/Donahue Gallery. \\
\textbf{Rationale selected by Baseline:} \\
The passage mentions that the Taunton Art Association was founded in 1973. The passage also mentions that Art Euphoric was founded in 2008. Therefore, Art Euphoric was founded after the Taunton Art Association. The passage does not mention when the Hughes/Donahue Gallery and the Trescott Street Gallery were founded. \\
\midrule
\textbf{Passage:} \\
In 1085, Guadalajara was retaken by the Christian forces of Alfonso VI . The chronicles say that the Christian army was led by Alvar Fanez de Minaya, one of the lieutenants of El Cid. From 1085 until the Battle of Las Navas de Tolosa in 1212, the city suffered wars against the Almoravid and the Almohad Empires. In spite of the wars, the Christian population could definitely settle down in the area thanks to the repopulation with people from the North who received their first fuero in 1133 from Alfonso VII. In 1219, the king Fernando III gave a new fuero to the city. During the reign of Alfonso X of Castile, the protection of the king allowed the city to develop its economy by protecting merchants and allowing markets. \\
\textbf{Question:} \\
How many years after the people of the North received their first fuero from Alfonso VII did king Fernando III give a new fuero to the city? \\
\textbf{Rationale selected by REPS:} \\
After the people of the North received their first fuero from Alfonso VII in 1133, Fernando III granted a new fuero to the city in 1219. Therefore, the city received its fuero 86 years after the first. \\
\textbf{Rationale selected by Baseline:} \\
The passage mentions that the first fuero was given in 1133 by Alfonso VII. By looking at the dates, it can be concluded that 287 years later, the people of the North received their first fuero. After this, 86 more years passed before Fernando III gave a new fuero to the city. Thus, it took 373 years from the time the people of the North received their first fuero to the time Fernando III gave a new fuero to the city. \\
\bottomrule
\end{tabular}
\caption{Examples of rationales selected by REPS and the baseline verifier for questions from DROP.}
\label{tab:example_drop_reps_baseline}
\end{table*}

\section{Examples of Rationales Selected by REPS and Baseline}
\label{sec:appendix-reps-baseline-examples}
Examples of rationales selected by REPS and the baseline verifier for the same question are provided across three datasets: ARC (Table \ref{tab:example_arc_reps_baseline}), DROP (Table \ref{tab:example_drop_reps_baseline}), and StrategyQA (Table \ref{tab:example_strategyqa_reps_baseline}).

\section{Human Annotation}
\label{sec:appendix-manual-annotation}
\paragraph{Validity Judgement}
To measure the alignment between human judgments and GPT-4's judgments, the authors conducted a validity assessment task. 
100 questions were randomly selected from the StrategyQA dataset. 
For each of these questions, the authors evaluated whether the generated rationale was factually accurate and logically consistent, given the supporting facts for each question. 
Each rationale was manually labeled as either valid or invalid based on these criteria.
The validity annotations obtained from GPT-4 show a Cohen's kappa of 0.62 with the authors' manual validity annotations.

\paragraph{Pairwise Win Rate Evaluation}
In the pairwise win rate evaluation, the authors compared 100 pairs of rationales selected by REPS and the baseline verifier for each dataset (ARC, DROP, StrategyQA). 
For ARC and StrategyQA, the evaluation of factuality was conducted with reference to supporting facts. 
We chose the rationale that was more factually accurate and logically consistent, with the rationales presented in a randomized order to mitigate bias.

\end{document}